\documentclass{article}

\usepackage{PRIMEarxiv}
\usepackage{tcolorbox}

\usepackage[utf8]{inputenc} % allow utf-8 input
\usepackage[T1]{fontenc}    % use 8-bit T1 fonts
\usepackage{hyperref}       % hyperlinks
\usepackage{url}            % simple URL typesetting
\usepackage{booktabs}       % professional-quality tables
\usepackage{amsfonts}       % blackboard math symbols
\usepackage{nicefrac}       % compact symbols for 1/2, etc.
\usepackage{microtype}      % microtypography
\usepackage{lipsum}
\usepackage{fancyhdr}       % header
\usepackage{graphicx}       % graphics
\graphicspath{{media/}}     % organize your images and other figures under media/ folder
\usepackage{tabularx}
\usepackage{caption}
\usepackage{amsmath}

\title{Benchmarking LLMs and SLMs for Patient Reported Outcomes
}

\author{
  Matteo Marengo \\
  ENS Paris-Saclay \\
  Gif-sur-Yvette\\
  \And
  Jarod Levy\\
  ENS Paris-Saclay \\
  Gif-sur-Yvette\\
   \And
  Jean-Emmanuel Bibault \\
  Hôpital Européen Georges Pompidou \\
  Paris\\
}

\begin{document}
\maketitle

\begin{abstract}
\hspace{1cm} LLMs have transformed the execution of numerous tasks, including those in the medical domain. Among these, summarizing patient-reported outcomes (PROs) into concise natural language reports is of particular interest to clinicians, as it enables them to focus on critical patient concerns and spend more time in meaningful discussions. While existing work with LLMs like GPT-4 has shown impressive results, real breakthroughs could arise from leveraging SLMs as they offer the advantage of being deployable locally, ensuring patient data privacy and compliance with healthcare regulations. This study benchmarks several SLMs against LLMs for summarizing patient-reported Q\&A forms in the context of radiotherapy. Using various metrics, we evaluate their precision and reliability. The findings highlight both the promise and limitations of SLMs for high-stakes medical tasks, fostering more efficient and privacy-preserving AI-driven healthcare solutions.
\end{abstract}

% keywords can be removed

\section{Introduction}
\hspace{1cm} Large Language Models (LLMs) have recently shown great potential in the medical field, accomplishing tasks like form summarization with impressive accuracy. Powerful models such as GPT-4 and Gemini are noted for their high performance. However, there is a significant privacy concern with medical data. To ensure complete patient privacy, these models should be operated locally within hospitals, rather than on external servers by using an API. One possible solution is to use small language models (SLMs). These models are advantageous because they can be used locally on a computer or within a server. The key question is: can these smaller, specialized medical models achieve the same level of accuracy as the larger, general models on simple medical tasks ? Indeed, medical tasks require a precision that is almost perfect to guarantee that patients will not suffer any side effects.

\begin{figure}[h!]
    \centering
    \includegraphics[width = \textwidth]{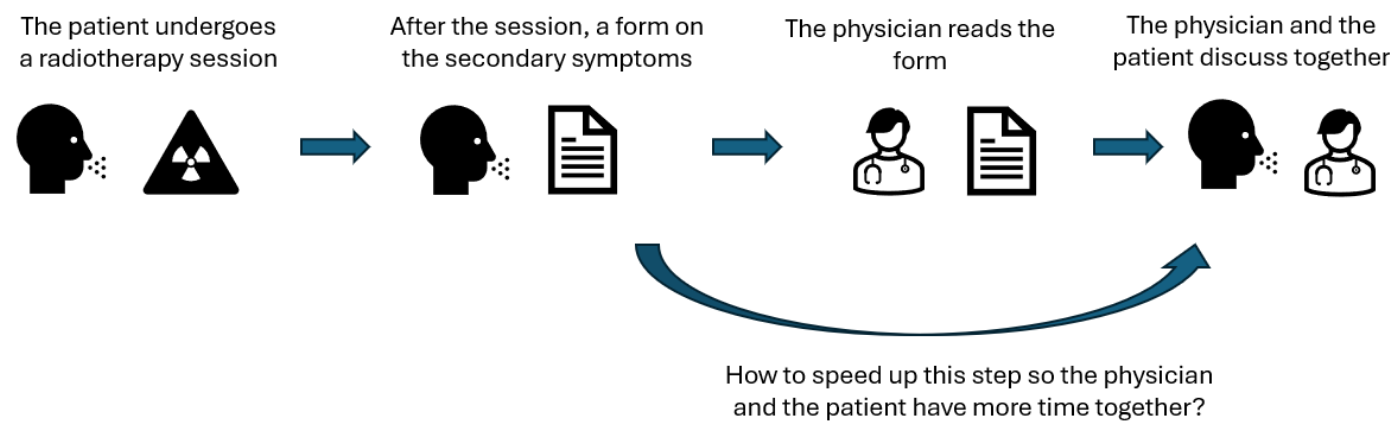}
    \caption{Sum-up of the task wanted by the radiotherapist. The classical procedure is first presented and then the enhancement being brought by LLMs.}
    \label{fig:sumupQAtask}
\end{figure}

The task that is going to be benchmarked in this study is the \textbf{patient reported outcomes summarization}. For patients undergoing cancer radiotherapy, managing and reporting side effects is crucial. Patients typically meet with their clinician after each treatment session, filling out forms to report any side effects they've experienced. These forms need to be summarized into a few lines of natural language to highlight the major symptoms, allowing clinicians to save time and focus on critical issues that might otherwise be overlooked. During these weekly on-treatment visits, which last about 10 to 20 minutes, it is essential to discuss any radiation toxicity symptoms with the radiation oncologist and address patient concerns. However, manual reporting often involves asking specific questions to screen for urinary and digestive toxicities related to radiation therapy. Studies have shown that physicians may underreport patient toxicity \cite{underestimate}, potentially compromising patient care. Additionally, the process of symptom gathering and documentation can be time-consuming for physicians, reducing the time available to address other important patient concerns. The typical process a patient follows during a radiotherapy session, and the step we aim to optimize is shown in Fig \ref{fig:sumupQAtask}.  \\

In our study, we will compare and assess small language models for this task. We have developed an evaluation method to measure their accuracy. Based on our findings, we will discuss how these models can be integrated into the current healthcare system. The paper is structured as follows: first, we present the related works that utilize large language models for medical tasks, with a particular focus on question-answering and summarization. Next, we introduce the methods and data used to evaluate the smaller language models. This is followed by the presentation of the results, and the paper concludes with a discussion on the applications of these language models in the medical field.

\section{Related Works}
\subsection{LLMs for Medical tasks}
\hspace{1cm} Recent work has shown \cite{JEB-GPT4} that using GPT-4 model achieves great results on the patient reported outcome task. This wonderful milestone in the medical field has seen many movements since. Indeed, many LLMs have seen the day, finetuned on data to answer to specific medical tasks and many of them claim to perform better than general LLMs. This is the case notably with Med-Gemini \cite{MEDGeminisaab2024capabilities} developed by Google which are languages fine-tuned on medical data. They achieve 91,1 \% accuracy on the MedQA benchmark \cite{MEDQABenchmark}. On other tasks, they show real progress compared to GPT-4 or their previous flagship Med-PaLM \cite{MEDPALM}. However, general language such as GPT-4 is nevertheless well considered for its abilities in the medical field \cite{nori2023capabilitiesGPT4}. For example, its performance at the United States Medical Licensing Examination (USMLE) was better than Med-PaLM. Some models such as BiomedGPT \cite{zhang2024biomedgpt} can even now use the vision module of GPT to treat biomedical images. One can also think of many LLMs based on BERT \cite{BERTpaper} architecture such as BioBERT \cite{BioBERT} that is pre-trained on a large corpus of biomedical texts (e.g PubMed abstracts, full-text articles). A nice review on these medical LLMs can be found here \cite{zhou2024medicalsurvey}. In addition to these LLMs performing efficiently on biomedical NLP tasks, some real usage of them have been done such in oncology \cite{RayaudOncology} or in otorhinolaryngology \cite{ORLLLM}.

\subsection{The Patient Reported Outcomes summarization task and SLMs}
\hspace{1cm} A significant paper addressing the task of summarizing key information from medical records \cite{adaptedLLMSoutperform} demonstrates that LLMs achieve either equivalent (45\%) or superior (36\%) results compared to those performed by medical experts. This study uses quantitative evaluation before conducting a clinical reader study. However, it's important to note that the evaluated tasks is still different than the task of patient reported outcome.
In addition, many LLMs have shown impressive results but in the context of the medical data total privacy has to be expected. Therefore, to run LLMs that get access to OpenAI servers for example is unlikely doable. Neither to run LLMs with more than 100 M parameters on the clinician laptop or the hospital server. This is why one focus would then be on small language models. Many of them have been developed such as Mistral models (especially Mistral7B \cite{jiang2023mistral7B}) or LLaMA models by Meta \cite{touvron2023llama}). They show great promise on many tasks. For example, Mistral 7B performs well on many benchmarks such as Math, Code or World Knowledge. This improvements is allowed thanks to strategies such as sliding window attention (SWA) mechanism ot Grouped-query attention (GQA). Furthermore, due to their reasonable size they can be easily finetuned for specific tasks. That is the case with BioMistral7B \cite{labrak2024biomistral}, a SLM that uses Mistral 7B as the fondation model an pre-trained on PubMed Central. SLMs that will be benchmarked are fine tuned for chat with instruction datasets, these models are called instruct ones.

\section{Methods}
\subsection{Data}

\hspace{1cm} The data that are being used for such an evaluation are the Patient-Reported Outcomes version of the Common Terminology Criteria for Adverse Events (PRO-CTCAE). It is a patient-reported outcome measurement system developed by the National Cancer Institute. The questions asked to the patients are for the vast majority not opened but a range of answers have to be selected. For our task, it has been decided to focus on 17 questions related to the Prostate Cancer. The detailed questions can be found in Appendix \ref{appendix 1}. Something important for the evaluation is that we associate to each question a keyword to later be able to evaluate the summary (e.g the keyword abdominal pain will be associated to the question "In the last 7 days, how OFTEN did you have PAIN IN THE ABDOMEN (BELLY AREA)? ").

\subsection{Models}
\hspace{1cm} To compare our small language models, we have decided to study State of the art models, therefore Mistral7B Instruct \cite{jiang2023mistral7B}, BioMistral7B \cite{labrak2024biomistral}, Llama2 7B Chat \cite{touvron2023llama} and Gemma 7b Instruct by Google will be extensively investigated. They will be compared to GPT-4. It is important to note that the models used here are instruct model meaning that they have been fine-tuned using specialized datasets that contain examples of instructions / responses making them more suited for tasks involving human interaction. 

\subsection{Prompt}
\hspace{1cm} The prompt is one of the key thing to consider when generating the summary as it could have a real impact on final results. The prompt that has been chosen in our case reflects and summarizes what a good sum-up should be. In addition, we pay attention to give the LLM an example of the expected output to guide the answer in the way we want. Our prompt for the evaluation will then be :

\begin{tcolorbox}[colback=gray!10, colframe=black, title=Prompt, fonttitle=\bfseries]
You are an experienced radiation oncologist physician. You are provided this list of questions and answers about patient symptoms during their weekly follow-up visit during radiotherapy.
Please summarize the following data into two sentences of natural language for your physician colleagues. Indicate in your summary only the most important symptoms using exactly the group of words in parenthesis at the end of the question.

A “yes” is important. Each time, include the number of radiation treatments and the answer of the last question if answered.
Provide the summarization in the English language.

\end{tcolorbox}

\begin{tcolorbox}[colback=gray!10, colframe=blue, title=Example of summarized symptoms, fonttitle=\bfseries]
This patient, after 30 radiation treatments, reports very severe symptoms including fatigue, flatulence and diarrhea. The patient also experiences severe skin burns from radiation and leakages that impact his daily activities. He also mentioned occasional fever as an additional symptom.

\end{tcolorbox}

\subsection{Metrics}
\hspace{1cm} To evaluate how well a language model performs on a specific task is challenging. Commonly used metrics, such as ROUGE or BLEU, rely on reference summaries for comparison, which is not applicable in our case (an extensive dataset of written summaries by medical doctors would be required). Our approach was to reduce the language model's freedom. Each MCQ question is associated with one or more keywords, and the prompt explicitly instructs the model to use these exact keywords when discussing the symptom.

\begin{table}[h]
\label{tab-questions}
\centering
\begin{tabular}{|c|c|}
\hline
\textbf{Question Type} & \textbf{Scale} \\
\hline
No | Yes & 0 | 1 \\
\hline
None | Mild | Moderate | Severe | Very Severe & 0 | 0.25 | 0.5 | 0.75 | 1 \\
\hline
Not at all | A little bit | Somewhat | Quite a bit |Very much & 0 | 0.25 | 0.5 | 0.75 | 1 \\
\hline
Never | Rarely | Occasionally | Frequently | Almost Constantly & 0 | 0.25 | 0.5 | 0.75 | 1 \\
\hline
\end{tabular}\\
\vspace{0.1cm}
\caption{Conversion scales for different question types}
\label{table:scales}
\end{table}

Additionally, since there are multiple options for each question, we can score each answer based on its severity. In tab \ref{table:scales} is the detailed combinations There will be four metrics to consider; customized gravity score, Recall, the Kappa Cohen Index and a LLM based score. 

\subsubsection{Customized Score, Recall \& Kappa Cohen Index}

\hspace{1cm} We evaluate the summarizer using keywords related to the question at hand. The first metric, called the Severity Metric, measures how many important symptoms (those graded above 0.5) are included in the final summary compared to all important symptoms.

\[
S = \frac{K_s}{K_p}
\]

Here, \( S \) is the severity metric, \( K_s \) is the number of severe keywords in the summary, and \( K_p \) is the number of severe keywords in the patient-reported outcome. This helps us understand the proportion of significant symptoms correctly captured.

Recall is another metric that checks for completeness by balancing against false negatives—it assesses whether any important symptoms were missed in the summary.

\[
\text{Recall} = \frac{K_s}{K_p + K_{fn}}
\]

Here, \( K_s \) is the number of severe keywords in the summary, \( K_p \) is the number of severe keywords in the patient-reported outcome, and \( K_{fn} \) is the number of severe keywords present in the patient-reported outcome but not included in the summary.

We also use the Cohen's Kappa Index to gauge the summarizer's effectiveness.

\[
\text{KCI} = {(P_o - P_e)}/{(1 - P_e)}
\]

Here, \( P_o \) is the observed agreement between the summarizer and a theoretical observer, and \( P_e \) is the expected agreement by chance.

This detailed computation ensures that the Cohen's Kappa Index accounts for the agreement occurring by chance, providing a more accurate measure of the summarizer's performance.

Although these three metrics—Severity Metric, Recall, and Cohen's Kappa Index—differ slightly, they collectively assess how effectively serious symptoms are included in the summary. Ensuring that no critical symptoms are overlooked is vital to prevent severe harm to the patient.

\subsubsection{GPT 4 grade}

\hspace{1cm}An extra metric that can be used is a LLM based one with GPT 4. GPT 4 grade is to get an overall idea of the summary quality, we can ask GPT 4 to grade the summary between 0 and 1 depending on our criteria. In addition to the prompt, the original patient reported outcomes is given. The prompt given to GPT 4 has been defined like this :

\begin{tcolorbox}[colback=gray!10, colframe=black, title=Prompt, fonttitle=\bfseries]
As an evaluator of LLM summarization accuracy, you are tasked with assessing the quality of the provided summary in relation to the original patient data.

Assign a score ranging from 0 to 1, where 0 indicates a completely inaccurate summary and 1 signifies a perfect summary.

Your evaluation should specifically focus on how well the summary encapsulates all significant symptoms reported by the patient, with a strong emphasis on accuracy and the inclusion of critical information.

Incorrect or misleading information in the summary should highly impact the score towards the lower end.

Your response should consist solely of the numerical score without additional commentary.
\end{tcolorbox}

\subsection{Evaluation pipeline}

\begin{figure}[h!]
    \centering
    \includegraphics[width=\linewidth]{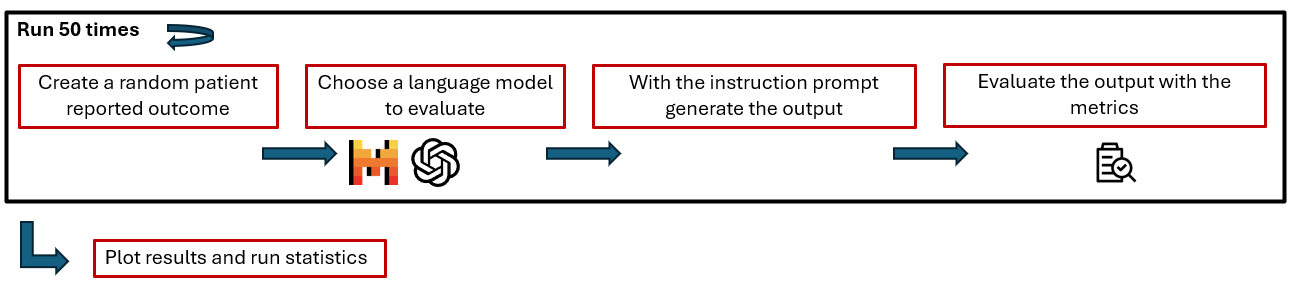}
    \caption{LLM Evaluation Pipeline. Presentation of the different steps to evaluate the LLMs and SLMs.}
    \label{fig:llm evaluation pipeline}
\end{figure}

\hspace{1cm}Our evaluation pipeline will proceed as follows. We will begin by instructing the model to focus only on identifying severe symptoms in patient responses, using the corresponding keywords associated with those symptoms. Each patient answer will be scored for severity, applying a threshold of 0.5 to determine when a symptom is considered severe. We will then assess the model's ability to include any applicable free-form responses provided by the patients. Additionally, we will score the number and relevance of treatments the model suggests in response to the identified severe symptoms. Finally, we will grade the overall summary generated by the model, evaluating it for clarity, accuracy, and completeness in capturing the patient's severe symptoms and recommended treatments. The pipeline is illustrated in Fig \ref{fig:llm evaluation pipeline}.

\subsection{Experimental conditions}

A random prompt generator has been created, which generates different answers to questions based on possible variations. Each time an LLM is benchmarked, 50 random prompts are created and assessed. The original weights of the LLMs were obtained from HuggingFace and computations were performed using an A100 GPU. We set the temperature to 0.7 and the maximum number of tokens to 200.

\section{Results}
\subsection{Qualitative evaluation}

\begin{figure}[h!]
    \centering
    \includegraphics[width=\textwidth]{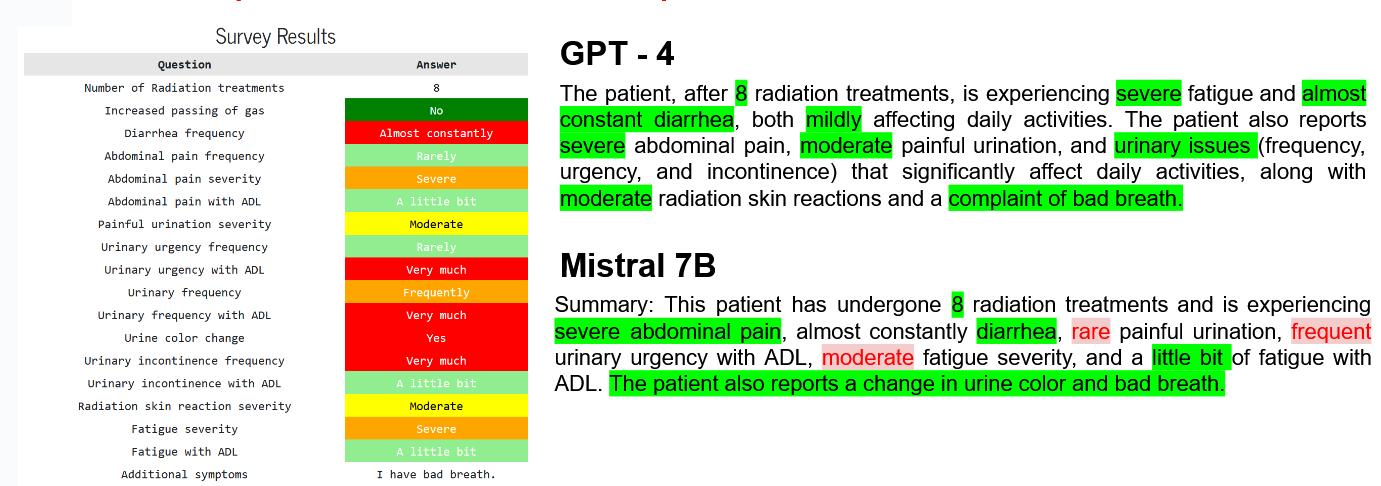}
    \caption{Qualitative performance comparison between GPT-4 and Mistral 7B. Left: symptoms with the related severity. Right: summary for both GPT-4 and Mistral 7B}
    \label{fig:perfor-comp}
\end{figure}

\hspace{1cm} First, we conduct a qualitative evaluation between GPT-4 and Mistral 7B Instruct using the same prompt. As shown in Figure~\ref{fig:perfor-comp}, Mistral 7B does not consider some symptoms or even underestimates them. For example, Mistral 7B classifies painful urination as rare instead of moderate, rates urinary urgency with ADL as frequent instead of very frequent, and assesses fatigue severity as moderate when it should be severe. This can lead to incorrect conclusions, as some severe symptoms are not taken into account by the model.

\subsection{Quantitative evaluation}

\hspace{1cm} The evaluation results for the four metrics across the five language models—GPT-4, Mistral-7B-Instruct, BioMistral-7B, Llama-2-7b-chat-hf, and Gemma-7b-it—reveal clear trends and insights about their performance. GPT-4 consistently outperforms all other models across all metrics, demonstrating both accuracy and reliability. It achieves the highest mean Severity Score (0.83) (cf Tab \ref{modelevolutation-severity}) with the lowest standard deviation (0.14), indicating both superior performance and consistency. Similarly, GPT-4 leads in Recall (mean: 0.56, std: 0.16) (cf Tab \ref{modelevolution-recall}), highlighting its ability to identify relevant true positives effectively. Its relatively high Kappa Cohen Index (mean: 0.34, std: 0.23) further reflects good agreement with the dataset’s ground truth, making it the most robust and reliable model overall.

\begin{table}[h!]
\centering
\begin{tabular}{|l|c|c|c|c|}
\hline
\textbf{Model} & \textbf{Mean Score} & \textbf{Std Score} & \textbf{Max Score} & \textbf{Min Score} \\ 
\hline
{GPT4} & \textbf{0.83} & \textbf{0.14} & {1.00} & {0.50} \\
\hline
{Mistral-7B-Instruct-v0.1} & 0.62 & 0.19 & 1.00 & 0.00 \\
\hline
{BioMistral-7B} & {0.55} & {0.24} & {1.00} & {0.00} \\
\hline
{Llama-2-7b-chat-hf} & {0.60} & {0.21} & {1.00} & {0.00} \\
\hline
{Gemma-7b-it} & {0.60} & {0.26} & {1.00} & {0.00} \\
\hline
\end{tabular}
\vspace{0.1 cm}
\caption{Models Evaluation - Severity Score}
\label{modelevolutation-severity}
\end{table}

\begin{table}[h!]
\centering
\begin{tabular}{|l|c|c|c|c|}
\hline
\textbf{Model} & \textbf{Mean Score} & \textbf{Std Score} & \textbf{Max Score} & \textbf{Min Score} \\ 
\hline
{GPT4} & {\textbf{0.56}} & {\textbf{0.16}} & {1.00} & {0.33} \\
\hline
{Mistral-7B-Instruct-v0.1} & 0.36 & 0.13 & 0.71 & 0.13 \\
\hline
{BioMistral-7B} & {0.32} & {0.21} & {1.00} & {0.00} \\
\hline
{Llama-2-7b-chat-hf} & {0.34} & {0.13} & {0.67} & {0.13} \\
\hline
{Gemma-7b-it} & {0.22} & {0.14} & {0.50} & {0.00} \\
\hline
\end{tabular}
\vspace{0.1 cm}
\caption{Models Evaluation - Recall}
\label{modelevolution-recall}
\end{table}

Among the SLMs, Mistral-7B-Instruct shows the best performance in Severity Score (mean: 0.62) and Recall (mean: 0.36). These results suggest its capability to identify true positives effectively, particularly when dealing with critical elements in the dataset. However, its performance drops significantly when evaluated on the Kappa Cohen Index (mean: -0.01), revealing a major inconsistency. This discrepancy suggests that while Mistral-7B-Instruct can identify many relevant instances, it also introduces substantial misclassifications, leading to poor overall agreement with the dataset’s labels.

Other SLMs, including BioMistral-7B, Llama-2-7b-chat-hf, and Gemma-7b-it, exhibit similar performance trends, with slightly lower Severity Scores (means ranging from 0.55 to 0.60) and Recall scores (means between 0.22 and 0.34). Their Kappa Cohen Index scores are also negative or near zero, indicating struggles with consistency and agreement with the ground truth. This suggests that while these models can identify some true positives, they lack the precision to align their predictions reliably with the dataset.

\begin{figure}[h!]
    \centering
    \begin{minipage}[b]{0.45\textwidth}
        \centering
        \includegraphics[width=\textwidth]{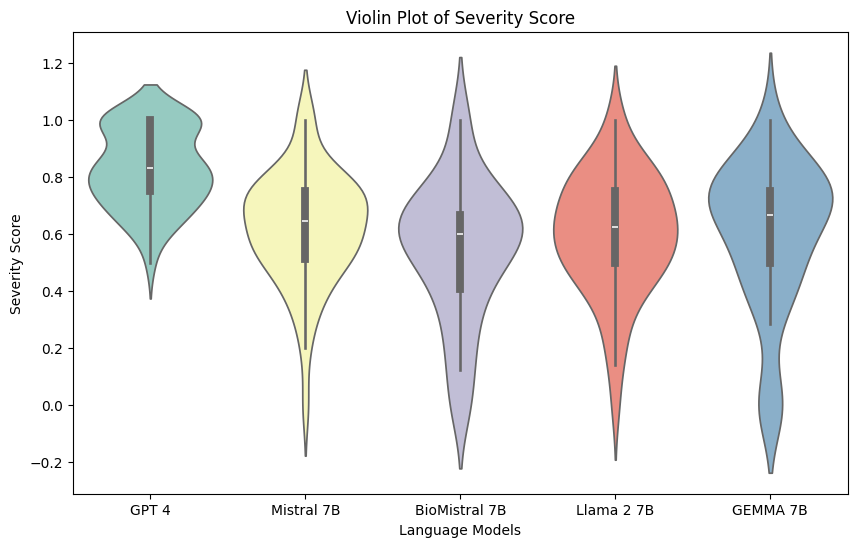}
        \caption{Violin Plot of the Severity Score - Comparison of the LMs}
        \label{fig:violinseverityscore}
    \end{minipage}
    \hfill
    \begin{minipage}[b]{0.45\textwidth}
        \centering
        \includegraphics[width=\textwidth]{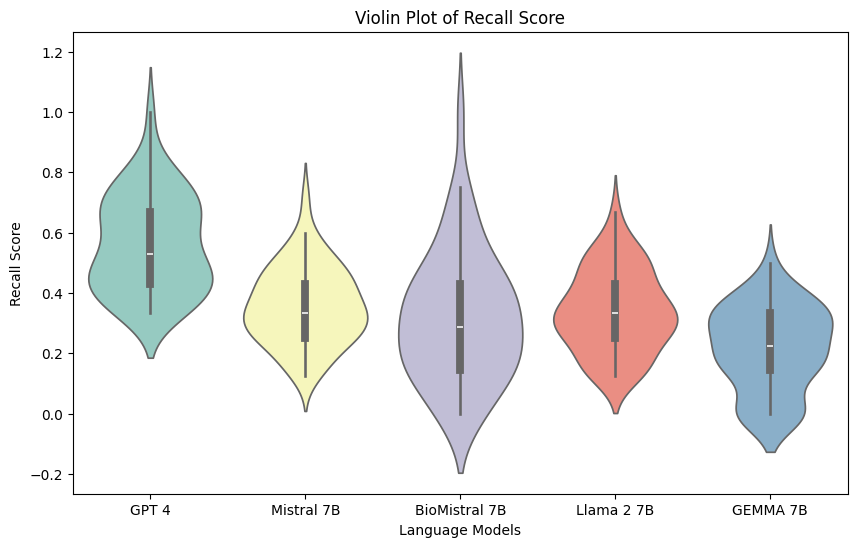}
        \caption{Violin Plot of the Recall Score - Comparison of the LMs}
        \label{fig:violinrecallscore}
    \end{minipage}
\end{figure}

The violin plots, Figure \ref{fig:violinseverityscore} and Figure \ref{fig:violinrecallscore}, further illustrate these findings. The Severity Score plot highlights GPT-4’s compact and high-performing distribution centered near 0.83, reflecting its consistent reliability. In contrast, Mistral-7B-Instruct exhibits a broader distribution, indicating higher variability in predictions, though it remains the best-performing SLM. The Recall plot mirrors this trend, with GPT-4’s concentrated distribution and Mistral-7B-Instruct standing out among the SLMs.

In conclusion, GPT-4 is the clear leader across all metrics, excelling in accuracy, consistency, and agreement with ground truth labels. Mistral-7B-Instruct, while promising in terms of Severity Score and Recall, struggles with misclassifications, as indicated by its poor Kappa Cohen Index. For practical applications, GPT-4 is the preferred model when accuracy and reliability are paramount. However, Mistral-7B-Instruct and other SLMs may still have value in tasks where identifying true positives (high Recall) is prioritized, despite challenges in maintaining agreement.

\begin{table}[h!]
\centering
\begin{tabular}{|l|c|c|c|c|}
\hline
\textbf{Model} & \textbf{Mean Score} & \textbf{Std Score} & \textbf{Max Score} & \textbf{Min Score} \\ 
\hline
{GPT4} & {0.34} & {0.23} & {1.00} & {-0.11} \\
\hline
{Mistral-7B-Instruct-v0.1} & -0.01 & 0.22 & 0.50 & -0.50 \\
\hline
{BioMistral-7B} & {0.05} & {0.29} & {0.64} & {-0.62} \\
\hline
{Llama-2-7b-chat-hf} & {0.11} & {0.22} & {0.53} & {-0.42} \\
\hline
{Gemma-7b-it} & {0.12} & {0.21} & {0.42} & {-0.35} \\
\hline
\end{tabular}
\vspace{0.1 cm}
\caption{Models Evaluation - Kappa Cohen Index}
\end{table}

\hspace{1cm} Concerning LLM based score (cf Tab \ref{llmbasedscore}), we observe interesting results. It shows that GPT 4 evaluates the response generated by GPT 4 as quasi perfect (0.97 mean). Furthermore, it also evaluates the other SLMs pretty good as there are all around 0.75. It proves once again that if GPT-4 remains the gold standard, Gemma-7b-it and Llama-2-7b-chat-hf offer promising alternatives among SLMs, particularly for applications where a balance between performance and computational efficiency is needed. Mistral-7B-Instruct and BioMistral-7B show potential but would benefit from further optimization to enhance their reliability and minimize prediction variability

\begin{table}[h!]
\centering
\begin{tabular}{|l|c|c|c|c|}
\hline
\textbf{Model} & \textbf{Mean Score} & \textbf{Std Score} & \textbf{Max Score} & \textbf{Min Score} \\ 
\hline
{GPT4} & {0.97} & {0.07} & {1.00} & {0.70} \\
\hline
{Mistral-7B-Instruct-v0.1} & 0.76 & 0.21 & 1.00 & 0.40 \\
\hline
{BioMistral-7B} & {0.74} & {0.22} & {1.00} & {0.20} \\
\hline
{Llama-2-7b-chat-hf} & {0.79} & {0.19} & {1.00} & {0.20} \\
\hline
{Gemma-7b-it} & {0.84} & {0.14} & {1.00} & {0.50} \\
\hline
\end{tabular}
\vspace{0.1 cm}
\caption{Models Evaluation - LLM Based Score}
\label{llmbasedscore}

\end{table}

\begin{figure}[h!]
    \centering
    \begin{minipage}[b]{0.45\textwidth}
        \centering
        \includegraphics[width=\textwidth]{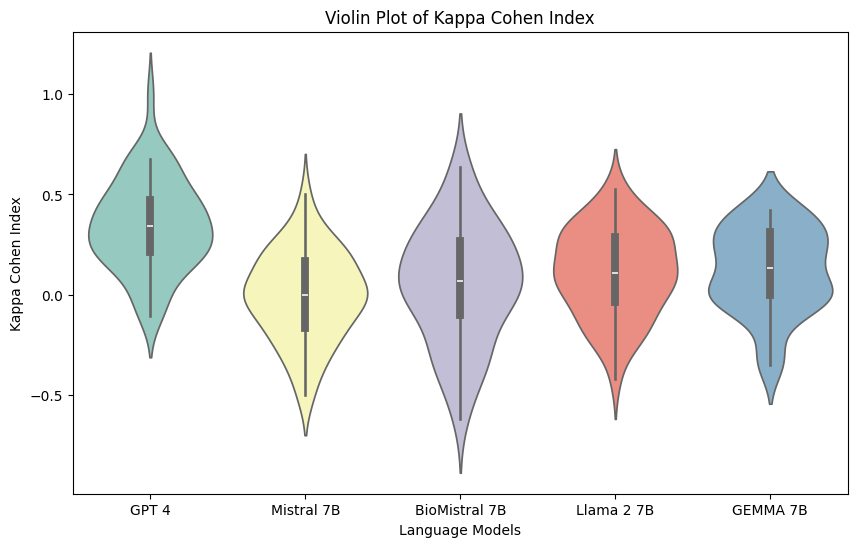}
        \caption{Violin Plot of the Kappa Cohen Index - Comparison of the LMs}
        \label{fig:kappacohenindex}
    \end{minipage}
    \hfill
    \begin{minipage}[b]{0.45\textwidth}
        \centering
        \includegraphics[width=\textwidth]{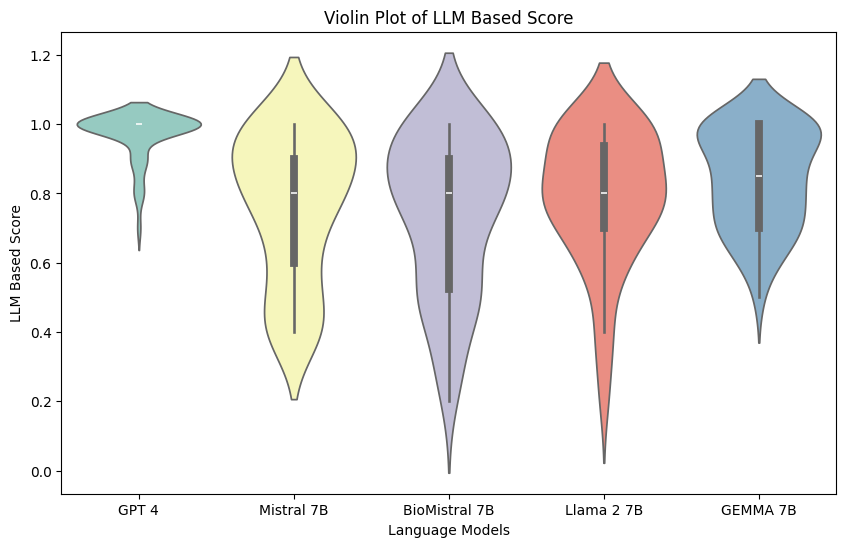}
        \caption{Violin Plot of the LLM based Score - Comparison of the LMs}
        \label{fig:violinllmbasedscore}
    \end{minipage}
\end{figure}

\section{Discussion}
\hspace{1cm} By benchmarking these Small Language Models against Large Language Models like GPT-4, we observe a consistent performance gap across all metrics. These SLMs, which are of similar size (around 7 billion parameters), demonstrate comparable performance levels, with no single model significantly outperforming the others. However, the difference compared to GPT-4 is remarkable, with a mean performance increase of approximately 25\%. Despite this gap, a reassuring observation is that these models do prioritize severe symptoms in the summaries, which is a critical requirement for medical applications.

This benchmarking raises several important points and future directions. First, it is worth considering whether \textbf{fine-tuning} these smaller models on this specific task could yield better results. One promising idea would be to construct a custom dataset consisting of question/answer pairs accompanied by GPT-4-generated summaries. However, the limited improvements observed in the performance of BioMistral-7B—despite its fine-tuning on medical-specific data suggest that the challenge may not lie solely in the dataset domain but also in the summarization capabilities of the model itself. This indicates that fine-tuning efforts should focus not only on medical data but also on techniques specifically tailored to enhancing summarization performance.

\textbf{Another crucial aspect to explore is the practical integration of these language models into healthcare workflows}. While LLMs like GPT-4 or the upcoming Gemini model appear robust enough for high-stakes tasks like medical summarization, the readiness of institutions, healthcare professionals, and patients for such a paradigm shift is equally important. Beyond technical performance, factors such as trust, interpretability, regulatory approval, and the ethical implications of AI-generated content must be thoroughly addressed. Ensuring that these models are transparent and their predictions are explainable is essential for widespread adoption in healthcare.

\section{Conclusion}
\hspace{1cm} In their current state, SLMs show promise but appear too fragile for reliable summarization tasks in critical medical contexts. However, their efficiency and scalability make them valuable candidates for low-stakes applications or as complementary tools to larger models. 

Looking ahead, the potential for LLMs such as GPT-4 and Gemini to revolutionize healthcare is immense. These models demonstrate consistent accuracy and reliability, making them strong candidates for tasks like summarization, patient communication, and clinical decision support. As the field evolves, combining the efficiency of SLMs with the robustness of LLMs through hybrid architectures or ensemble methods could offer a balanced approach, addressing both scalability and accuracy challenges. Ultimately, the integration of language models into healthcare requires a multidisciplinary effort, blending technical advancements with rigorous validation and human-centered design to ensure their safe and effective deployment.

%Bibliography
\bibliographystyle{unsrt}  
\bibliography{references}  

\newpage
\section{Appendix}
\subsection*{Questions selected and their answers}\label{appendix 1}

\begin{table}[h!]
\centering
\caption{Questions, Possible Answers, and Keywords}
\label{tab:questions_answers_keywords}
\begin{tabularx}{\textwidth}{|>{\scriptsize}X|X|X|}
\hline
\textbf{Question} & \textbf{Possible Answers} & \textbf{Keywords} \\ \hline
How many radiation treatments have you had? & open & N/A \\ \hline
In the last 7 days, what was the SEVERITY of your FATIGUE, TIREDNESS, OR LACK OF ENERGY at its WORST? & None, Mild, Moderate, Severe, Very severe & FATIGUE \\ \hline
In the last 7 days, how much did FATIGUE, TIREDNESS, OR LACK OF ENERGY INTERFERE with your usual or daily activities? & Not at all, A little bit, Somewhat, Quite a bit & FATIGUE \\ \hline
In the last 7 days, did you have any INCREASED PASSING OF GAS (FLATULENCE)? & Yes, No & FLATULENCE \\ \hline
In the last 7 days, how OFTEN did you have LOOSE OR WATERY STOOLS (DIARRHEA)? & Never, Rarely, Occasionally, Frequently, Almost constantly & DIARRHEA \\ \hline
In the last 7 days, how OFTEN did you have PAIN IN THE ABDOMEN (BELLY AREA)? & Never, Rarely, Occasionally, Frequently, Almost constantly & ABDOMINAL ABDOMEN PAIN \\ \hline
In the last 7 days, what was the SEVERITY of your PAIN IN THE ABDOMEN (BELLY AREA) at its WORST? & None, Mild, Moderate, Severe, Very severe & ABDOMINAL ABDOMEN PAIN \\ \hline
In the last 7 days, how much did PAIN IN THE ABDOMEN (BELLY AREA) INTERFERE with your usual or daily activities? & Not at all, A little bit, Somewhat, Quite a bit, Very much & ABDOMINAL ABDOMEN PAIN \\ \hline
In the last 7 days, what was the SEVERITY of your PAIN OR BURNING WITH URINATION at its WORST? & None, Mild, Moderate, Severe, Very severe & URINATION PAIN \\ \hline
In the last 7 days, how OFTEN did you feel an URGE TO URINATE ALL OF A SUDDEN? & Never, Rarely, Occasionally, Frequently, Almost constantly & URGE URGES URINATE \\ \hline
In the last 7 days, how much did SUDDEN URGES TO URINATE INTERFERE with your usual or daily activities? & Not at all, A little bit, Somewhat, Quite a bit, Very much & URGE URGES URINATE \\ \hline
In the last 7 days, were there times when you had to URINATE FREQUENTLY? & Never, Rarely, Occasionally, Frequently, Almost constantly & FREQUENT FREQUENCY URINATION \\ \hline
In the last 7 days, how much did FREQUENT URINATION INTERFERE with your usual or daily activities? & Not at all, A little bit, Somewhat, Quite a bit, Very much & URINATION URINE FREQUENCY \\ \hline
In the last 7 days, did you have any URINE COLOR CHANGE? & Yes, No & URINE COLOR COLORATION \\ \hline
In the last 7 days, how OFTEN did you have LOSS OF CONTROL OF URINE (LEAKAGE)? & Never, Rarely, Occasionally, Frequently, Very much, Almost constantly & LEAKAGE \\ \hline
In the last 7 days, how much did LOSS OF CONTROL OF URINE (LEAKAGE) INTERFERE with your usual or daily activities? & Not at all, A little bit, Somewhat, Quite a bit, Very much & LEAKAGE \\ \hline
In the last 7 days, what was the SEVERITY of your SKIN BURNS FROM RADIATION at their WORST? & None, Mild, Moderate, Severe, Very severe, Not applicable & SKIN BURNS \\ \hline
Finally, do you have any other symptoms that you wish to report? & open & N/A \\ \hline
Summarization Language & English, French, Portuguese & N/A \\ \hline
\end{tabularx}
\end{table}

\end{document}